\documentclass[journal]{IEEEtran}

\usepackage{multirow}
\usepackage{times}
\usepackage{epsfig}
\usepackage{graphicx}
\usepackage{amsmath}
\usepackage{amssymb}
\usepackage{xcolor}
\usepackage{subfigure}
\usepackage{url}

\usepackage{algorithm}
\usepackage{subfigure}

\begin{document}
%
\title{Unsupervised Cross-Domain Recognition by Identifying Compact Joint Subspaces}

\author{Yuewei~Lin,~\IEEEmembership{Student~Member,~IEEE,}
        Jing~Chen,~\IEEEmembership{Member,~IEEE,}
        Yu~Cao,~\IEEEmembership{Member,~IEEE,}
        Youjie~Zhou,~\IEEEmembership{Student~Member,~IEEE,}
        Lingfeng~Zhang,~\IEEEmembership{Student~Member,~IEEE,}
        Yuan~Yan~Tang,~\IEEEmembership{Fellow,~IEEE,}
        and~Song~Wang,~\IEEEmembership{Senior~Member,~IEEE}
\thanks{This work was supported by AFOSR FA9550-11-1-0327, NSF IIS-1017199. This work was also supported by the Research Grants of University of Macau MYRG2015-00049-FST, MYRG2015-00050-FST, RDG009/FST-TYY/2012, and the Science and Technology Development Fund (FDCT) of Macau 100-2012-A3, 026-2013-A. This research project was also supported by the National Natural Science Foundation of China 61273244.}
\thanks{Y.~Lin, Y.~Zhou and S.~Wang are with the Department of Computer Science and Engineering, University of South Carolina, Columbia, SC 29208 USA (e-mail: ywlin.cq@gmail.com; zhou42@email.sc.edu; songwang@cec.sc.edu).}
\thanks{J. Chen is with Chongqing University, Chongqing, China (e-mail: chenjingmc@gmail.com).}
\thanks{Y. Cao is with IBM Almaden Research Center, 650 Harry Road, San Jose, CA 95120 USA (e-mail: caoy@us.ibm.com).}
\thanks{L. Zhang is with University of Houston, Houston, TX 77004, USA (e-mail: lzhang34@uh.edu).}
\thanks{Y. Y. Tang is with University of Macau, Macau, China (e-mail: yytang@umac.mo).}
\thanks{Manuscript submitted July 8, 2015.}}

\markboth{Submitted to IEEE Transactions on Cybernetics}%
{Shell \MakeLowercase{\textit{et al.}}: Bare Demo of IEEEtran.cls for Journals}

\maketitle

\begin{abstract}
This paper introduces a new method to solve the cross-domain recognition problem. Different from the traditional domain adaption methods which rely on a global domain shift for all classes between source and target domain, the proposed method is more flexible to capture individual class variations across domains. By adopting a natural and widely used assumption -- ``the data samples from the same class should lay on a low-dimensional subspace, even if they come from different domains'', the proposed method circumvents the limitation of the global domain shift, and solves the cross-domain recognition by finding the compact joint subspaces of source and target domain. Specifically, given labeled samples in source domain, we construct subspaces for each of the classes. Then we construct subspaces in the target domain, called anchor subspaces, by collecting unlabeled samples that are close to each other and highly likely all fall into the same class. The corresponding class label is then assigned by minimizing a cost function which reflects the overlap and topological structure consistency between subspaces across source and target domains, and within anchor subspaces, respectively. We further combine the anchor subspaces to corresponding source subspaces to construct the compact joint subspaces. Subsequently, one-vs-rest SVM classifiers are trained in the compact joint subspaces and applied to unlabeled data in the target domain. We evaluate the proposed method on two widely used datasets: object recognition dataset for computer vision tasks, and sentiment classification dataset for natural language processing tasks. Comparison results demonstrate that the proposed method outperforms the comparison methods on both datasets.
\end{abstract}


\IEEEpeerreviewmaketitle

\section{Introduction}
\label{sec:Introduction}
\begin{figure}[htbp]
\centering
\includegraphics[width=\columnwidth]{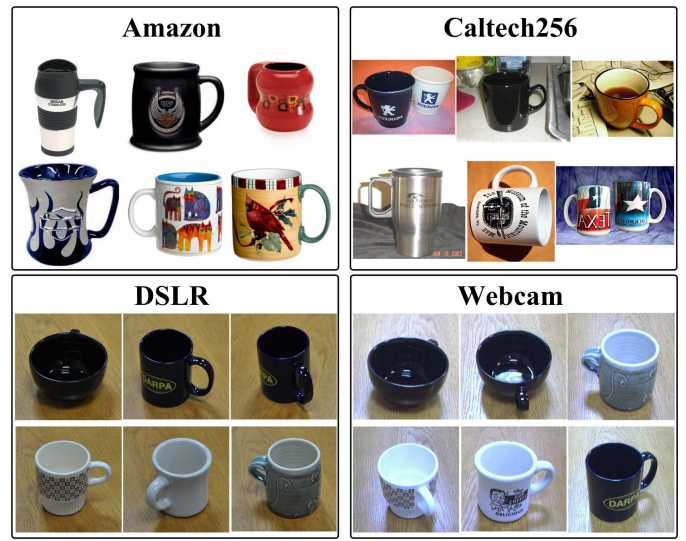}
\caption{Sample images from four different domains: Amazon, Caltech256, DSLR and Webcam.}
\label{fig:domain_examples}
\end{figure}

\IEEEPARstart{M}{any} machine learning methods often assume that the training data (labeled) and testing data (unlabeled) are from the same feature space and following similar distributions. However this assumption may not be true in many real applications. Namely the training data is obtained from one domain, while the testing data is coming from a different domain. As a visual example, Figure~\ref{fig:domain_examples} shows coffee-mug images collected from four different domains (Amazon, Caltech256, DSLR and Webcam), which present different image resolutions (Webcam vs DSLR), viewpoints (Webcam vs Amazon), background complexities (Amazon vs Caltech256) and object layout patterns, etc. 

\begin{figure*}[htbp]
\centering
\subfigure[]{\includegraphics[width=0.8\columnwidth]{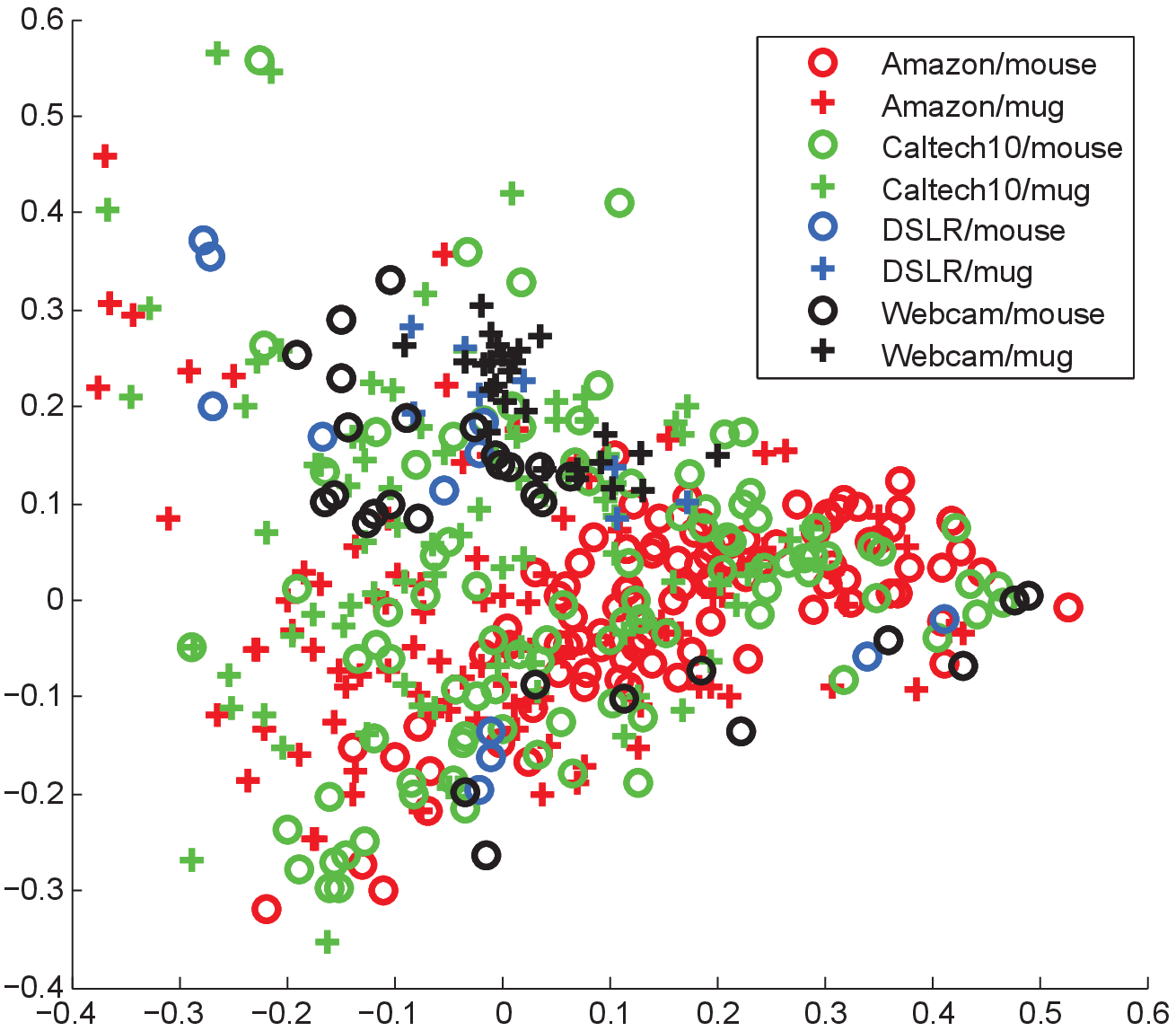}}
\hspace{0.3in}
\subfigure[]{\includegraphics[width=0.8\columnwidth]{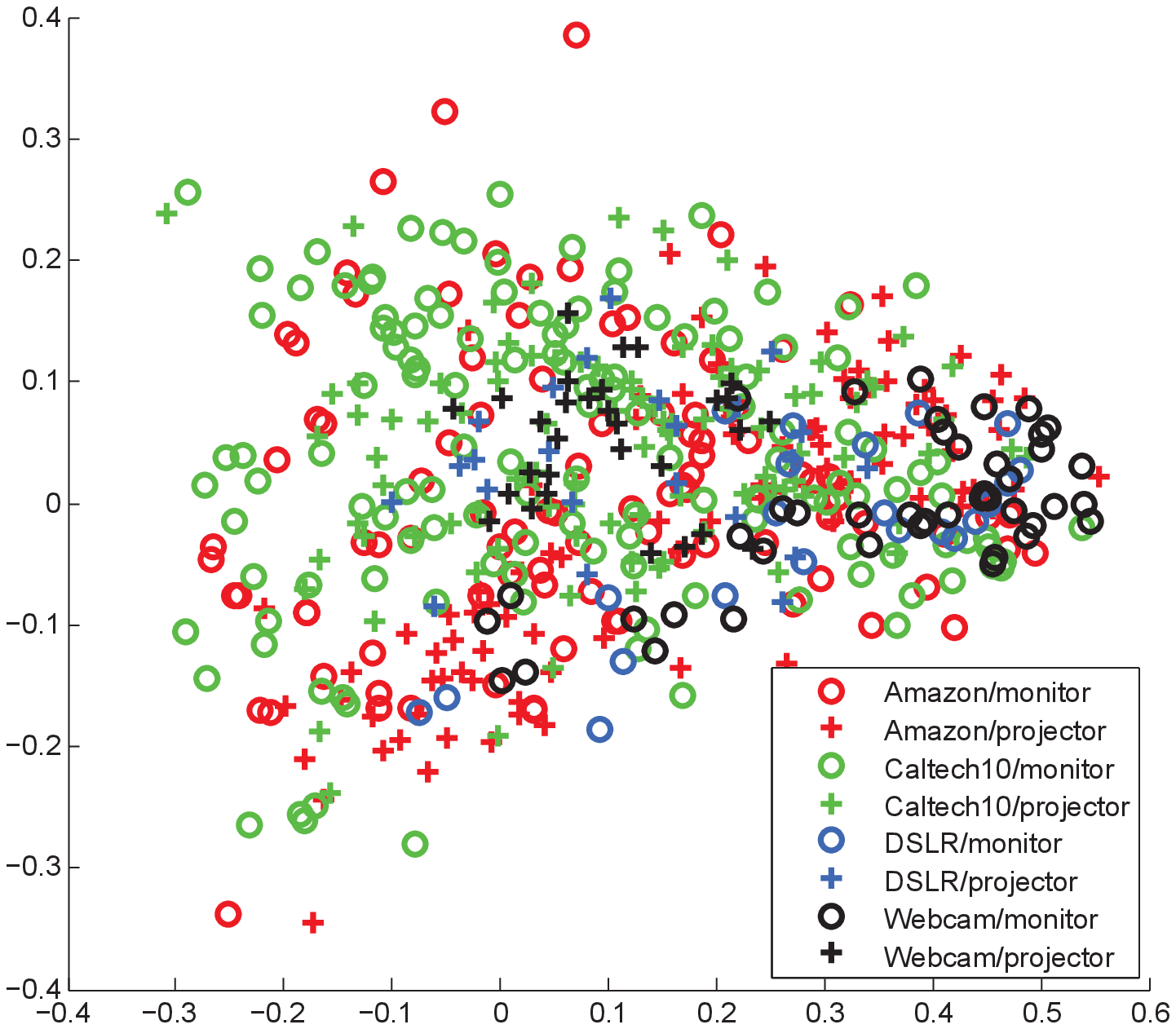}}
\caption{Illustrations of sample distributions of different domains in feature space. The 2-D plots are the first two feature dimensions reduced from original 800 dimensional SURF feature space (describe in Section~\ref{sec:Object_recognition}), by using PCA. (a) The distributions of ``mouse" and ``mug" in four domains. (b) The distributions of ``monitor" and ``projector" in four domains. Best viewed in color.}
\label{fig:distributions}
\end{figure*}

On the other hand, the samples also show different distributions in the feature space, as illustrated in Figure~\ref{fig:distributions}. The 2-D plots are the first two feature dimensions reduced from original 800 dimensional SURF feature space (described in Section~\ref{sec:Object_recognition}), by using PCA. Figure~\ref{fig:distributions}~(a) and (b) show the distributions of ``mouse" and ``mug", and ``monitor" and ``projector" in different domains, respectively. It is clear to see that the samples from different domains have different distributions. Moreover, the relations between two classes in different domains are also different. Taking Figure~\ref{fig:distributions}~(a) for example, in ``Webcam" domain, the mug samples (black crosses) usually locate at the right-top side of the mouse ones (black circles); but in ``Amazon" domain, the mug samples (red crosses) usually locate at the left side of the mouse ones (red circles).

These domain differences lead to a dilemma that 1) directly applying the classifiers trained from one domain to another may result in significant degraded performance~\cite{Torralba11}; 2) labeling data in each domain as training samples would be very expensive, especially in large-scale applications. The dilemma consequently poses the cross-domain recognition problem, namely how to utilize the labeled data in a \textit{source} domain to classify/recognize the unlabeled data in a \textit{target} domain.


To achieve cross-domain recognition, a number of \emph{Domain Adaption} (DA) methods have been developed to adapt the classifier from one domain to another~\cite{Pan10}. The subspace based DA has been found to be very effective to handle cross-domain problem~\cite{Gopalan11,Gong12,Jhuo12,Ni13,Shekhar13,Cui14,Long14}. They either constructed a set of intermediate subspaces for modeling the shifts between domains~\cite{Gopalan11,Gong12,Ni13,Cui14}, or generated a domain-invariant subspace in which the data from source and target domains can represent each other well~\cite{Jhuo12,Shekhar13,Long14}. All these methods mentioned above utilize the data from each domain all together to generate a single subspace for each domain. In practice, however, the intrinsic feature shift of each class may not be exactly the same. The existing methods can obtain a global domain shift, but ignore the individual class difference across domains.

\begin{figure}[htbp]
\centering
\includegraphics[width=\columnwidth]{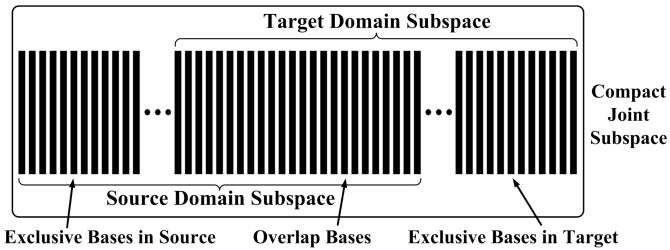}
\caption{An illustration of a compact joint subspace between source and
	target domains for a specific class. This compact joint subspace
	consists of overlap bases between domains, which represent the
	intrinsic characteristics of this class implicitly, and exclusive
	bases of different domains, which represent the exclusive characteristics
	for each domain.}
	\label{fig:bases}
\end{figure}

To circumvent the limitation of the global domain shift, we adopt a natural and widely used assumption that ``the data samples from the same class should lay on a low-dimensional subspace, even if they come from different domains~\cite{Elhamifar13}''. This assumption is not only held on many computer vision tasks, such as face recognition under varying illumination~\cite{Basri03} and handwritten digit recognition~\cite{Hastie98}, but also used as a human cognitive mechanism for visual object recognition~\cite{Dicarlo12}. Note that this assumption does not mean that the target data samples exactly lay on the subspace of the source samples, since different domains show subspace shift~\cite{Ni13}. Figure~\ref{fig:bases} gives an illustration of a compact joint subspace covering source domain and target domain for a specific class. The source and target subspaces have the overlap which implicitly represents the intrinsic characteristics of the considered class. They have their own exclusive bases because of the domain shift, such as the varying illumination or changing the view perspectives.

Based on the above assumption, we propose a new method that solves the cross-domain recognition by finding the compact joint subspaces of source and target domain. Specifically, given labeled samples in source domain, we construct subspaces for each of the classes. Then we construct subspaces in the target domain, called \textit{anchor subspaces}, by collecting unlabeled samples that are close to each other and highly likely all fall into the same class. The corresponding class label is then assigned by minimizing a cost function which reflects the overlap and and topological structure consistency between subspaces across source and target domains, and within anchor subspaces, respectively. We further combine the anchor subspace to corresponding source subspaces to construct the compact joint subspaces for each class. Subsequently, the SVM classifier is trained by using the samples in the compact joint subspace and applied to the unlabeled data in the target domain for classification.

The contributions of this paper are: \textbf{1)} by assuming that the data samples from one specific class, even though they come from different domains, should lay on a low dimensional subspace, we generate one compact joint subspace for each class independently. Each compact joint subspace carries the information not only about the intrinsic characteristics of the corresponding class, but also about the specificity for each domain. \textbf{2)} To construct the compact joint subspaces, we first generate anchor subspaces in the target domain, assign labels to them, and combine these anchor subspaces to the corresponding source subspaces. \textbf{3)} We propose a cost function that implicitly maximizes the overlap between source subspace and target subspace for each class as well as maintains the topological structure in the target domain. We use principal angles as the subspace distances in the cost function instead of data-sample distances that were usually used in previous methods.

Note that proposed method does not need to get the orthogonal bases for the subspaces; instead, we use the data samples themselves as the over-complete bases to represent the subspace implicitly.

The remainder of this paper is organized as follows. We elaborate related works in Section~\ref{sec:related_work}. Section~\ref{sec:Model} describes the proposed method. Quantitative experimental results are demonstrated in Section~\ref{sec:Result}. Section~\ref{sec:conclusion} concludes the paper.

\section{Related Work}\label{sec:related_work}
For the cross-domain recognition problem, Domain Adaptation is the most closely related work which is known as a type of fundamental methods in machine learning and computer vision. Here, we give a brief review of this topic. Please refer to~\cite{Patel15} for a comprehensive survey. 

The traditional DA algorithms can be categorized into two types, i.e., (semi-)supervised domain adaptation and unsupervised domain adaptation, based on the availability of labeled data from the target domain. (Semi-)supervised DA assumes that there are some labeled data available in the target domain. Daume~\cite{Daume07} proposed to map the data from both source and target domains to a high dimensional feature space, and trained the classifiers in this new feature space. Saenko et al. \cite{Saenko10} proposed a metric learning approach that can adapt labeled data of few classes from the target domain to the unlabeled target classes. In \cite{Kumar10}, the authors proposed a co-regularization model that augmented feature space to jointly model source and target domains. In \cite{Chen11}, Chen et al. proposed a co-training based domain adaptation method. They first train an initial category model on samples from source domain and then use it for labeling samples from target domain. The category model keeps updating by the newly labeled target samples through co-training. Pan and his colleagues \cite{Pan11} analyzed the transfer component that maps both of domains on kernel space to preserve some properties of domain-specific data distributions. Duan et al. \cite{Duan12} proposed a SVM based method which minimized the mismatches between source and target domains, using both labeled and unlabeled data. Shekhar et al. \cite{Shekhar13} proposed to learn a single dictionary to represent both source and target domains. The work in \cite{Donahue13} proposed to use a linear transformation map features from the target domain to the source domain and generate the classification model training on the source domain and target samples based on feature transformation. Motivated by the recent success of deep learning, some hierarchical domain adaptation methods are also proposed ~\cite{Donahue14}, which need the large scale data to (pre-)train the deep neural network model. In \cite{Xiao15}, Xiao et al. proposed a semi-supervised kernel matching based domain adaptation method that learns a prediction function on the source domain while mapping the target samples to similar source samples by matching the target kernel matrix to the source kernel matrix.

Unsupervised DA, on the other hand, does not use any label information in the target domain, which is also considered as more challenging and more useful in real-world applications. In \cite{Gopalan11,Gopalan14}, Gopalan et al. constructed a set of intermediate subspaces along the geodesic path that links the source and target domains on the Grassmann manifold. In \cite{Gong12}, Gong et al. proposed a geodesic flow kernel to model shift between the source and target domains. In \cite{Ni13}, a new intermediate subspaces construction method was proposed, which constructed the subspaces by gradually reducing the reconstruction error of the target data instead of using the manifold walking strategies. Jhuo et al. \cite{Jhuo12} learned a transformation so that the source samples can be represented by target samples in a low-rank way. Fernando et al. proposed to learn a mapping function which aligns the sample representations from source and target domains \cite{Fernando13}. Tommasi et al. \cite{Tommasi13} proposed a naive Bayes’ nearest neighbor-based domain adaptation algorithm that iteratively learns  a class metric while inducing a large margin separation among classes for each sample. Baktashmotlagh et al.~\cite{Baktashmotlagh14} proposed to use the Riemannian metric as a measure of distance between the distributions of source and target domain. In \cite{Long14}, Long et al. proposed to learn a domain invariant representation by jointly performing the feature matching and instance weighting. Cui et al.~\cite{Cui14} treated samples from each domain as one points (i.e., covariance matrices) on a Riemannian manifold, and then interpolate some intermediate points along the geodesic, which are used to bridge the two domains.

The algorithm proposed in this paper is an unsupervised cross-domain recognition method. It is different from the traditional domain adaption methods due to it constructs the low dimensional compact joint subspaces for each class independently, which will avoid the global domain shift limitation and capture the individually domain variations for each class.

\section{Proposed Model}\label{sec:Model}

\subsection{Problem}\label{sec:Problem}
Suppose there are two sets of data samples, one from source domain $\mathcal{S}$, denoted as $\{X^S_i\}_{i=1}^{N_S}\in R^{d\times N_S}$, and the other one from target domain $\mathcal{T}$, denoted as $\{X^T_i\}_{i=1}^{N_T}\in R^{d\times N_T}$, where $d$ is the data dimension, $N_S$ and $N_T$ denote the number of data samples in source and target domains, respectively. The labels of all data samples in source domain, denoted as $Y^S=\{y^S_i\}^{N_S}_{i=1}\in R^{C\times N_S}$, are known, where $C$ is the number of classes, $y^S_i\in \{0,1\}^C$ is a $C$ bit binary code of the $i$th data sample in source domain. If this data sample belongs to class $j$, the $j$th bit of $y^S_i$ is 1 and all other bits are 0. Our aim is to estimate $Y^T \in R^{C\times N_T}$, the labels of all the data samples in the target domain.

\subsection{Overview of the proposed method}
\label{sec:assumption_framework}

\begin{figure}[htbp]
\centering
\includegraphics[height=6.6cm]{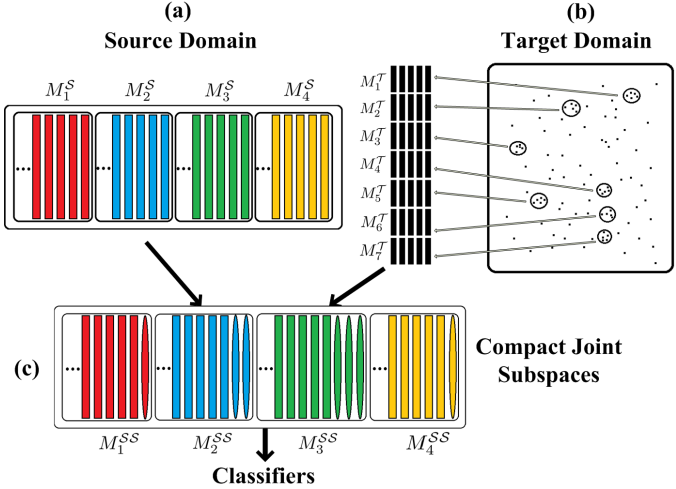}
\caption{The overview of the proposed model. (a). Subspaces for each class in source domain. The bars with the same color denote the bases of one class.(b). Anchor subspaces construction. The points in target domain are the data samples. Data samples in each circle denote a core subgroup and they construct an anchor subspace as one row of black bars. (c). Compact joint subspaces construction. The ellipses denote the bases from anchor subspaces (target domain). Best viewed in color.}
\label{fig:Framework}
\end{figure}

%

The proposed algorithm aims to construct a set of compact joint subspaces $\{M^{\mathcal{SS}}_i\}_{i=1}^{C}$, which cover source and target domains, one for each of $C$ classes, and then train the classifiers on these compact joint subspaces. As shown in Figure~\ref{fig:bases}, since a compact joint subspace is constituted by a source subspace and a target subspace, we need to construct the source and target subspaces first. Hence, the proposed algorithm consists of five steps:

\emph{\textbf{Constructing subspaces in source domain.}} We construct a set of subspaces $\{M_i^{\mathcal{S}}\}^C_{i=1}$, one for each class in the source domain. As mentioned above, we do not care about how to get a set of orthogonal bases to represent a subspace. For each class, we simply construct a subspace by using over-complete bases, i.e., take all the data that belong to this subspace. Hence, each source subspace $M_i^{\mathcal{S}}=\{ X_j^S|_{X_j^S\in C_i}\}$, where $C_i$ denotes the $i$th class, illustrated in Figure.~\ref{fig:Framework}(a).


\emph{\textbf{Constructing anchor subspaces in target domain.}}
To estimate the target subspace for each class, we construct a number of anchor subspaces in the target domain, denoted as $\{M_i^{\mathcal{T}}\}^{K}_{i=1}$, as illustrated in Figure.~\ref{fig:Framework}(b). These anchor subspaces are expected: 1) carry the information of target exclusive bases and 2) construct target subspace compactly. The data samples in target domain are naturally satisfy the first expectation. To satisfy the second expectation, the data samples in one anchor subspaces should be as much as possible from the same class, since the subspace constructed by samples from different class is usually less compact than the one constructed by samples from the same class. Thus, the basic idea is to ensure that each anchor subspace only contains data samples from a single class such that it can be combined to a source subspace for constructing the joint subspace. Since the data in the target domain are unlabeled, we construct anchor subspaces by grouping target data samples with high similarities. This is motivated by the \emph{locality principle} -- a data sample usually lies in close proximity to a small number of samples from the same class~\cite{Zografos13}.

\emph{\textbf{Labeling the anchor subspaces.}} Since the compact joint subspaces are constructed independently for each class, we assign a label for each anchor subspace. For this purpose, we propose a cost function that reflects 1) the cross-domain distance between the anchor subspace and the corresponding source subspaces; and 2) the within-domain topological relation of the anchor subspaces in the target domain. Shorter cross-domain distance between the corresponding subspaces actually reflects the desirability of more common bases in constructing a compact joint subspace. Therefore, the minimization of the proposed cost function implicitly reflects the maximization of the overlap between source and target subspace.


\emph{\textbf{Constructing compact joint subspaces.}} We construct compact joint subspace $M_i^{\mathcal{SS}}=\{ M_i^{\mathcal{S}}, M_j^{\mathcal{T}}|_{M_j^{\mathcal{T}}\in C_i}\}$, where $C_i$ denotes the $i$th class, as illustrated in Figure.~\ref{fig:Framework}(c). As mentioned before, we simply take all the data samples from all the involved subspaces as the over-complete bases in constructing the joint subspaces.

\emph{\textbf{Training classifiers on the compact joint subspaces.}} We train one-vs-rest linear SVM classifiers for each class using the labeled data in the compact joint subspace. And then we apply the linear SVM classifier to the unlabeled data in the target domain for classification.

\subsection{Anchor subspaces obtained in target domain}
\label{sec:anchor_subspace}
We construct each anchor subspace by selecting one target data sample and combining it with its nearest neighbors. This way, the obtained compact group of data samples are likely to be from the same class~\cite{Zografos13}. Specifically, we first apply the $K$-means algorithm to cluster all the target data into a large number of $Z$ groups. We set $Z=\frac{N_T}{\gamma}$, where $\gamma$ is the desired average group size. In each group of data, we find a compact \emph{core subgroup} consisting of a small number of $N$ samples, e.g., $N=5$, which are taken for constructing an anchor subspace. In this paper, the core subgroup for the group $L$ is constructed by following two steps.
\begin{enumerate}
\item Estimate the center of the core subgroup by finding the data sample $x^*$ to
\begin{equation}
\min_{x\in L}\sum_{y\in\mathcal{N}_{(N-1)}(x)}\|x-y\|_2,
\label{eq:compact}
\end{equation}
where $\mathcal{N}_{(N-1)}(x)$ denotes the $N-1$ nearest neighbors of $x$ in $L$.
\item Take $x^*\cup\mathcal{N}_{(N-1)}(x^*)$ as the core subgroup for constructing an anchor subspace.
\end{enumerate}
For the groups that contain less than $N$ data samples, we do not construct an anchor subspace.

\subsection{Labeling each anchor subspace}
\label{sec:assigning}
Note that, we have constructed $C$ subspaces in the source domain, one for each class, denoted as $\{M_i^{\mathcal{S}}\}^C_{i=1}$. Their corresponding labels, denoted as $Y=\{y_i\}_{i=1}^C\in R^{C\times C}$, which is an identity matrix, i.e., the $i$th bit of $y_i$ is 1 and all the other bits of $y_i$ are 0. In this section, we developed a new strategy to assign class labels $Y'=\{y'_i\}_{i=1}^K\in R^{C\times K}$ for the $K$ anchor subspaces $\{M_i^{\mathcal{T}}\}^{K}_{i=1}$ constructed in target domain. This strategy includes two main components: the similarity between subspaces and the cost function for subspace label assignment.


\subsubsection{Distance between subspaces}
\label{sec:Distance}
To calculate the distance between two subspaces, principal angles are usually used~\cite{Gong12,Hamm08}. Principle angles between subspaces, which often define between two orthonormal subspaces, serve as a classic tool in many areas of in computer science, such as computer vision and machine learning.

We follow the definition in~\cite{Hamm08}, given two orthonormal matrices $M_1$ and $M_2$, the principal angles
$0\leq\theta_1\leq\cdots\theta_m\leq\pi/2$ between two subspaces $span(M_1)$ and $span(M_2)$ is defined by:
\begin{equation}
	\begin{split}
		&\cos\theta_k = \max_{\bf{u_k\in span(M_1)}} \max_{\bf{v_k\in span(M_2)}} \bf{u'_k v_k}\\
		s.t.\,\,&\bf{u'_k u_k} = 1, \bf{v'_k v_k} = 1, \\
		&\bf{u'_k u_i} = 0, \bf{v'_k v_i} = 0, (i=1,\cdots,k-1).
	\end{split}
\end{equation}
The principal angles are related to the geodesic distance between $M_1$ and $M_2$ as $\sum_i\theta^2_i$ ~\cite{Hamm08}.

In practice, the principal angles are usually computed from the singular value decomposition (SVD) of $M'_1M_2$, i.e., $M'_1M_2 = U(\cos\Theta)V'$, where $U=[\bf{u_1,\cdots,u_m}]$, $V=[\bf{v_1,\cdots,v_n}]$, and $\cos\Theta$ is the diagonal matrix $diag\left(\cos\theta_1 \cdots cos\theta_{min(m,n)}\right)$.

In this paper, as shown in the later Section~\ref{sec:Cost_function}, we need both cross-domain subspace distance and within-domain subspace distance to define the cost function for anchor subspaces labeling. For the cross-domain subspace distance, e.g. between subspace $M_i^{\mathcal{S}}$ with $s_i$ data samples in the source domain and $M_j^{\mathcal{T}}$ with $t_j$ data samples in the target domain, we first orthogonalize both of them to obtain $\mathcal{M}_i^{\mathcal{S}}$ and $\mathcal{M}_j^{\mathcal{T}}$, and then calculate the distance as: 
$$D(M_i^{\mathcal{S}},M_j^{\mathcal{T}})\triangleq\sum_{i=1}^{min(t_i,s_j)}\sin\theta_i,$$
where $\theta_i$ come from the SVD of $\left(\mathcal{M}_i^{\mathcal{S}}\right)'\mathcal{M}_j^{\mathcal{T}}$
that $\left(\mathcal{M}_i^{\mathcal{S}}\right)'\mathcal{M}_j^{\mathcal{T}}=U(\cos\Theta)V'$.

Similarly, for the within-domain subspace distance, e.g., between two target (anchor) subspaces $M_i^{\mathcal{T}}$ and $M_j^{\mathcal{T}}$ that both of them have $N$ data samples we defined in Section~\ref{sec:assigning}, we first orthogonalize both of them to get $\mathcal{M}_i^{\mathcal{T}}$ and $\mathcal{M}_j^{\mathcal{T}}$, and then calculate the distance as:
$$D(M_i^{\mathcal{T}},M_j^{\mathcal{T}})\triangleq\sum_{i=1}^{N}\sin\theta'_i,$$
where $\theta'_i$ come from the SVD of $\left(\mathcal{M}_i^{\mathcal{T}}\right)'\mathcal{M}_j^{\mathcal{T}}$ that $\left(\mathcal{M}_i^{\mathcal{T}}\right)'\mathcal{M}_j^{\mathcal{T}}=U(\cos\Theta)V'$.

The above defined subspace distance follows the assumption that the samples within the same class share the same subspace even though they are from different domains. Consequently, the distance between subspaces across source and target domains of a specific class trends to smaller than that between different classes. We will show quantitative comparison in Section~\ref{sec:Distance} to demonstrate this advantage.

We further generate two affinity matrices, $C\times K$ matrix $A^{ST}$ to reflect the distances between $K$ anchor subspaces and $C$ source subspaces, and $K\times K$ matrix $A^{TT}$ to reflect the pairwise distances among $K$ anchor subspaces. More specifically, we have $A^{ST}(i,j)=\exp\left(-\frac{D(M^\mathcal{S}_i,M^\mathcal{T}_j)}{2\sigma^2}\right)$ and $A^{TT}(i,j)=\exp\left(-\frac{D(M^\mathcal{T}_i,M^\mathcal{T}_j)}{2\sigma^2}\right)$.

\subsubsection{Cost function and optimization}\label{sec:Cost_function}
Two important issues are considered in assigning a label to each anchor subspace: 1) the distance between an anchor subspace and the same-label source subspace should be small, and 2) the local topological structures in the target domain should be preserved \cite{Zhai13}, i.e., anchor subspaces with shorter distance are more preferable to be assigned to the same class. Considering these two issues, we propose the cost function as follow:
\begin{equation}
\mathcal{C}(Y')=\sum_{i=1}^C\sum_{j=1}^K\|y_i-y'_j\|^2 A^{ST}_{ij}+\rho\sum_{j=1}^K \sum_{j'=1}^K\|y'_j-y'_j\|^2 A^{TT}_{jj'}
\label{eq:costFunction_L2}
\end{equation}
where $A^{ST}$ and $A^{TT}$ are the affinity matrices of inter-domain subspace pairs and subspace pairs within target domain, respectively.

Adding a constant term $\sum_{i=1}^C\sum_{j=1}^C\|y_i-y_j\|^2 I_{ij}$ into $\mathcal{C}$ and splitting the first term into two parts, we get:
\begin{equation}
\begin{split}
\mathcal{C}(Y')=&\frac{1}{2}\sum_{i=1}^C\sum_{j=1}^K\|y_i-y'_j\|^2 A^{ST}_{ij}+\frac{1}{2}\sum_{j=1}^C\sum_{i=1}^K\|y_j-y'_i\|^2 A^{ST}_{ij}\\
&+\rho\sum_{j=1}^K \sum_{j'=1}^K\|y'_i-y'_j\|^2 A^{TT}_{jj'}+\sum_{i=1}^C\sum_{j=1}^C\|y_i-y_j\|^2 I_{ij},
\label{eq:costFunction_long}
\end{split}
\end{equation}
where $I_{ij}$ is the $ij$th element of the identity matrix $I$.

Note that the first and second terms are equal. The cost function $\mathcal{C}$ can be further written as:
\begin{equation}
\mathcal{C}(Y')=\sum_{i=1}^{C+K}\sum_{j=1}^{C+K}\|\mathcal{Y}_i-\mathcal{Y}_j\|^2 \mathcal{A}_{ij}, \,
s.t.\,\,\mathcal{Y}^\mathrm{T}\mathbf{1}=\mathbf{1}.
\label{eq:costFunction_final}
\end{equation}
where $\mathcal{Y}=[Y, Y']$, $\mathcal{A}=\left[\begin{array}{cc} \,\,\,\,\,\,I \,\,\,\,\,\,\,\,\,\,\,\,\,\,\,\,\,\,\,\,\,\,\,\, \frac{1}{2}A^{ST}\\ \frac{1}{2}\left(A^{ST}\right)^\mathrm{T} \,\,\,\,\,\, \rho A^{TT}\\ \end{array}\right]$. We also relax the constraint to this cost function by only requiring
the sum of each row in $\mathcal{Y}$ to be 1.


By including the constraint term, the cost function can be written in a matrix form \cite{Belkin03}
\begin{equation}
\mathcal{L}(\mathcal{Y},\lambda)=Tr\left(\mathcal{Y}\Delta\mathcal{Y}^\mathrm{T}\right)
+\lambda^\mathrm{T}\left(\mathcal{Y}^\mathrm{T}\mathbf{1}-\mathbf{1}\right)+\frac{\mu}{2}\|\mathcal{Y}^\mathrm{T}\mathbf{1}-\mathbf{1}\|^2_2,
\end{equation}
where $\Delta=\mathcal{D}-\mathcal{C}$ is the \emph{Laplacian matrix} of $\mathcal{A}$. $\mathcal{D}$ is the \emph{degree matrix} which is a diagonal matrix with $\mathcal{D}_{ii}=\sum_j\mathcal{A}_{ij}$. $\lambda\in\mathcal{R}^{C+K}$ is the Lagrange multiplier. To minimize the objective function $\mathcal{L}$, we separate it into two steps to update its two unknowns $\mathcal{Y}$ and $\lambda$ alternately, using the following two steps:

\textbf{Step 1:} Having $\lambda$ fixed, optimize $\mathcal{Y}$ by computing the derivative of $\mathcal{L}$ with the respect to $\mathcal{Y}$ and setting it to be zero:
\begin{equation}
\frac{\partial\mathcal{L}(\mathcal{Y},\lambda)}{\partial\mathcal{Y}}=0
\Rightarrow \mathcal{Y}\Delta+\mathbf{1}\lambda^\mathrm{T}+
\mu\left(\mathbf{1}\mathbf{1}^\mathrm{T}\mathcal{Y}-\mathbf{1}\mathbf{1}^\mathrm{T}\right)=0.
\label{eq:step1}
\end{equation}

Note that $\mathcal{Y}$ contains two parts $Y$ and $Y'$, with $Y$ is known. Thus, to solve it, as in \cite{Zhu03}, we first split the Laplacian matrix $\Delta$ into 4 blocks along the $C$th row and column:
\begin{eqnarray*}
\Delta=\left[
\begin{array}{c c}
\Delta_{CC} & \Delta_{CK}\\
\Delta_{KC} & \Delta_{KK}
\end{array}
\right].
\end{eqnarray*}

Similarly, we separate $\lambda$ into two parts:
\begin{displaymath}
\lambda_1=[\lambda_1,\lambda_2,\cdots,\lambda_C]^\mathrm{T} \mbox{ and } \lambda_2=[\lambda_{C+1},\lambda_{C+2},\cdots,\lambda_{C+K}]^\mathrm{T}.
\end{displaymath}
Then $Y'$ can be updated by solving the following equation:
\begin{equation}
Y'^{(k+1)}\Delta_{KK}+\mu\mathbf{1}\mathbf{1}^\mathrm{T}Y'^{(k+1)}=\mu\mathbf{1}\mathbf{1}^\mathrm{T}
-Y\Delta_{CK}-\mathbf{1}\lambda_2^{(k)\mathrm{T}}.
\label{eq:solution}
\end{equation}
The solution is given by the Lyapunov equation \cite{matrixcodebook}. With this solution, $\mathcal{Y}^{(k+1)}$ can be
achieved by putting $Y$ and $Y'^{(k+1)}$ together as $\mathcal{Y}^{(k+1)}=[Y, Y'^{(k+1)}]$.

\textbf{Step 2:} Having $\mathcal{Y}$ fixed,  perform a gradient ascending update with the step of $\mu$ on Lagrange multipliers as:
\begin{equation}
\lambda^{(k+1)}=\lambda^{(k)}+\mu\left(\mathcal{Y}^{(k+1)^\mathrm{T}}\mathbf{1}-\mathbf{1}\right).
\label{eq:step2}
\end{equation}

To initialize this optimization process, we simply set $\lambda^{(0)}$, $Y'^{(0)}$ to be zero, and set the maximal number of iteration $maxIter$ to be 10000. This whole algorithm is summarized in Algorithm.~\ref{alg:Assign}.

\begin{algorithm}
\textcolor{black}{ \caption{Labeling the Anchor Subspaces.\label{alg:Assign}}}

\textbf{\textcolor{black}{Input:}}\textcolor{black}{\ Affinity matrix $\mathcal{A}$, $maxIter$, labels $Y$ for $M{^\mathcal{S}}$}{\par}
\textcolor{black}{}\textbf{Output:}{\ Labels $Y'$ for $M^{\mathcal{T}}$.}\\
\\
\textcolor{black}{\textbf{Initialization:} $\lambda^{(0)}$, $Y'^{(0)}$ to zero}{\par}
\textcolor{black}{\ \ 1: \textbf{while}}\ \ (not converged $\&$ not achieve the maxIter){\par}
\textcolor{black}{\ \ 2:\ \ \ \ Update the $Y'^{(k+1)}$ by solving the following equation:}{\par}
\vspace{-0.1in}
\begin{small}
\begin{equation*}
\ \ \ \ \ \ \ \ \ \ Y'^{(k+1)}\Delta_{KK}+\mu\mathbf{1}\mathbf{1}^\mathrm{T}Y'^{(k+1)}=\mu\mathbf{1}\mathbf{1}^\mathrm{T}
-Y\Delta_{CK}-\mathbf{1}\lambda_2^{(k)\mathrm{T}}
\end{equation*}
\end{small}
\textcolor{black}{\ 3:\ \ \ \ Update $\mathcal{Y}^{(k+1)}=[Y, Y'^{(k+1)}]$.}{\par}
\textcolor{black}{\ \ 4:\ \ \ \ Update $\lambda^{(k+1)}=\lambda^{(k)}+\mu\left(\mathcal{Y}^{(k+1)^\mathrm{T}}\mathbf{1}-\mathbf{1}\right)$.}{\par}
\textcolor{black}{\ \ 5:\ \ \ \ Check the convergence.}{\par}
\textcolor{black}{\ \ 6:\ \ \textbf{end while}}{\par}
\end{algorithm}

Since we only require the sum of each row in $\mathcal{Y}$ to be 1, after we get $Y'$, we set the bit with the maximal value in each row to 1 and all the other bits to 0.

\section{Experimental Results}\label{sec:Result}
In this section, we first give the evaluation results on the subspace distance that we used in the proposed algorithm to demonstrate that this distance is suitable for our algorithm. Then we evaluate the proposed algorithm comprehensively on two widely used cross domain recognition datasets: object recognition image dataset for computer vision tasks and sentiment classification dataset for natural language processing tasks. 

\subsection{Evaluation on the distance we used in proposed algorithm}\label{sec:Distance}

The distance matrix in Table~\ref{tab:distanceMatrix} is given to show that the subspace based distance is suitable for our method. In this table, we use the data from object recognition dataset (details in Section \ref{sec:Object_recognition}). Each column denotes the distance between a specific class (C1,$\cdots$,C10) in source domain and a class (C1,$\cdots$,C10) in target domain. Note that all the numbers are the average of all 12 pairs of the source and target domain (describe in Section~\ref{sec:Single_cases}). We can see that the distances, across two domains, between the same class  are relatively smaller than those between different classes. Thus, the results also demonstrate the assumption that the samples with the same class share the same subspace even though they are from different domains, i.e., the distance between subspaces across source and target domains of a specific class trends to be smaller than that between different classes.

\begin{table}[htbp]
\tabcolsep 3pt \caption{The distance matrix between classes across source and target domains.}
\begin{center}
\renewcommand{\arraystretch}{1.3}
\def\temptablewidth{1\columnwidth} {\rule{\temptablewidth}{1.5pt}}
\begin{tabular*}{\temptablewidth}{@{\extracolsep{\fill}}c|cccccccccc}
    &\parbox[c]{.2cm}{C1}    &C2    &C3     &C4     &C5    &C6     &C7     &C8     &C9    &C10\\
\hline
C1  &\bf{0.86}  &0.92  &0.92  &0.91  &0.91  &0.90  &0.90  &0.91  &0.91  &0.91\\
\hline
C2  &0.92  &\bf{0.85}  &0.92  &0.90  &0.94  &0.92  &0.93  &0.92  &0.93  &0.94\\
\hline
C3  &0.92  &0.92  &\bf{0.85}  &0.92  &0.90  &0.90  &0.91  &0.92  &0.92  &0.92\\
\hline
C4  &0.91  &0.90  &0.92  &\bf{0.87}  &0.92  &0.90  &0.90  &0.91  &0.91  &0.91\\
\hline
C5  &0.91  &0.94  &0.90  &0.92  &\bf{0.86}  &0.90  &0.91  &0.92  &0.92  &0.92\\
\hline
C6  &0.90  &0.92  &0.90  &0.90  &0.90  &\bf{0.86}  &0.88  &0.90  &0.92  &0.90\\
\hline
C7  &0.90  &0.93  &0.91  &0.90  &0.91  &0.88  &\bf{0.87}  &0.90  &0.92  &0.90\\
\hline
C8  &0.91  &0.92  &0.92  &0.91  &0.92  &0.90  &0.90  &\bf{0.91}  &0.92  &0.92\\
\hline
C9  &0.91  &0.93  &0.92  &0.91  &0.92  &0.92  &0.92  &0.92  &\bf{0.91}  &0.92\\
\hline
C10 &0.91  &0.94  &0.92  &0.91  &0.92  &0.90  &0.90  &0.92  &0.92  &\bf{0.90}\\
\hline
\end{tabular*}
{\rule{\temptablewidth}{1.5pt}}
\end{center}
\label{tab:distanceMatrix}
\end{table}

\subsection{Cross-domain recognition on object dataset}\label{sec:Object_recognition}
The first dataset that we evaluate on is an image dataset. The whole dataset has four sub-datasets, which we use as four domains, with 2533 images from 10 classes in total. The first three sub-datasets were collected by \cite{Saenko10}, which include images from amazon.com (Amazon), collected with a digital SLR (DSLR) and a webcam (Webcam). The fourth domain is Caltech-256 dataset (Caltech) \cite{Caltech256}. Following the way of feature extraction for each image in \cite{Ni13}, we first use a SURF \cite{Bay08} detector to extract points of interest from each image. We then randomly select a subset of the points of interest and quantize their descriptors to $800$ visual words using the $K$-means clustering. Finally, we construct a 800-dimensional feature vector for each image using the bag-of-visual-words technique. For simplicity, hereafter we use ``A", ``C", ``D" and ``W" to denote the ``Amazon", ``Caltech", ``DSLR" and ``Webcam" domains, respectively.

\subsubsection{Single source domain and single target domain}\label{sec:Single_cases}

\begin{table*}[ht]
\caption{Results of single source and target domain on the object recognition dataset. ``-" denotes that there is no resualt reported before.}
\begin{center}
\renewcommand{\arraystretch}{1.3}
\def\temptablewidth{0.7\textwidth} {\rule{\temptablewidth}{1.5pt}}
\begin{tabular*}{\temptablewidth}{@{\extracolsep{\fill}}c||c|c|c|c|c|c}
Model                 &C$\rightarrow$A  &C$\rightarrow$W  &C$\rightarrow$D  &A$\rightarrow$C  &A$\rightarrow$W &A$\rightarrow$D\\ 
\hline
K-SVD \cite{Aharon06}  &20.5$\pm$0.8     & -               &19.8$\pm$1.0     &20.2$\pm$0.9 	  &16.9$\pm$1.0     & -          \\
SGF \cite{Gopalan11}   &48.9$\pm$0.7     &42.9$\pm$0.8     &44.0$\pm$1.0     &40.0$\pm$0.3    &35.0$\pm$0.7     &34.9$\pm$0.6\\
GFK \cite{Gong12}      &40.4$\pm$0.7     &35.8$\pm$1.0     &41.1$\pm$1.3     &37.9$\pm$0.4	  &35.7$\pm$0.9     &35.2$\pm$0.9\\
Metric \cite{Saenko10} &33.7$\pm$0.8     & -               &35.0$\pm$1.1     &36.0$\pm$1.0    &21.7$\pm$0.5     & -          \\
ITL \cite{Shi12}       &49.2$\pm$0.6     &43.3$\pm$0.7     &44.4$\pm$1.2     &38.5$\pm$0.4    &40.0$\pm$1.3     &39.6$\pm$0.6\\
SI \cite{Ni13}         &45.4$\pm$0.3     &37.0$\pm$5.1     &42.3$\pm$0.4     &40.4$\pm$0.5    &37.9$\pm$0.9     &32.1$\pm$4.5\\
SA(SVM) \cite{Fernando13}  &46.1         &38.9             &39.4             &39.9            &39.6             &38.8       \\
DASC \cite{Cui14}      &49.8$\pm$0.4     &45.4$\pm$0.9     &48.5$\pm$0.8     &39.1$\pm$0.3    &37.7$\pm$0.7     &39.3$\pm$0.8\\
TJM \cite{Long14}      &46.8             &39.0             &44.6             &39.5            &42.0             &45.2\\
CJS (ours)        &\bf{59.1$\pm$1.2} &\bf{52.2$\pm$2.6} &\bf{53.0$\pm$3.5} &\bf{47.6$\pm$1.1} &\bf{42.2$\pm$2.9} &\bf{47.9$\pm$2.2} \\
\hline
\hline
Model                 &W$\rightarrow$C  &W$\rightarrow$A  &W$\rightarrow$D &D$\rightarrow$C  &D$\rightarrow$A &D$\rightarrow$W\\
\hline
K-SVD \cite{Aharon06}   &13.2$\pm$0.6    &14.2$\pm$0.7    & -               & -              &14.3$\pm$0.3    &46.8$\pm$0.8 \\
SGF \cite{Gopalan11}    &32.3$\pm$0.4    &35.1$\pm$0.5    &72.9$\pm$0.7     &34.9$\pm$0.3    &34.7$\pm$0.4    &82.0$\pm$0.6 \\
GFK \cite{Gong12}       &29.3$\pm$0.4    &35.5$\pm$0.7    &71.2$\pm$0.9     &32.7$\pm$0.4    &36.1$\pm$0.4    &79.1$\pm$0.7 \\
Metric \cite{Saenko10}  &32.3$\pm$0.8    &38.6$\pm$0.8    & -               & -              &30.3$\pm$0.8    &55.6$\pm$0.7 \\
ITL \cite{Shi12}        &32.2$\pm$0.3    &35.2$\pm$0.3    &75.6$\pm$0.8     &34.7$\pm$0.3    &39.6$\pm$0.4    &83.6$\pm$0.5 \\
SI \cite{Ni13}          &\bf{36.3$\pm$0.3} &38.3$\pm$0.3  &79.5$\pm$2.0     &\bf{35.5$\pm$1.8} &39.1$\pm$0.5  &86.2$\pm$1.0 \\
SA(SVM) \cite{Fernando13} &31.8          &39.3            &77.9             &35.0            &\bf{42.0}       &82.3  \\
DASC \cite{Cui14}       &33.3$\pm$0.3    &36.3$\pm$0.4    &71.2$\pm$0.9     &32.7$\pm$0.4    &36.5$\pm$0.3    &88.3$\pm$0.4 \\
TJM \cite{Long14}       &30.2            &30.0            &89.2             &31.4            &32.8            &85.4 \\
CJS (ours)             &33.5$\pm$1.6  &\bf{39.5$\pm$1.3}  &\bf{89.4$\pm$1.8} &34.5$\pm$1.9  &37.9$\pm$1.6  &\bf{89.3$\pm$1.7} \\
\end{tabular*}
{\rule{\temptablewidth}{1.5pt}}
\end{center}
\label{tab:single}
\end{table*}


We report the results on all twelve possible pairs of source- and target-domain combinations, followed. We ran our algorithm 20 times for each object-recognition task and gave the average accuracy rate ($\%$) and standard deviation ($\%$) in Table.~\ref{tab:single}. We compare our algorithm with nine other methods, including K-SVD~\cite{Aharon06}, SGF~\cite{Gopalan11}, GFK~\cite{Gong12}, Metric~\cite{Saenko10}, ITL~\cite{Shi12}, SI~\cite{Ni13}, SA~\cite{Fernando13}, DASC~\cite{Cui14} and TJM~\cite{Long14}. Their results in Table.~\ref{tab:single} are obtained from previous papers, mostly by the original authors. It can be seen that our algorithm performs best in 9 out 12 domain pairs. In particular, in four domain pairs our algorithm significantly outperforms (by more than $5\%$) than all the comparison methods, i.e., C$\rightarrow$A, C$\rightarrow$D, A$\rightarrow$C and C$\rightarrow$W. Our algorithm shows a comparable performance with the best performed method in the other three domain pairs. Note that the ``Metric" method~\cite{Saenko10} is a semi-supervised method.

\subsubsection{Multiple source/target domains}\label{sec:Multiple_cases}

We then evaluate the performance when there are multiple source/target domains. To get the fair comparison with other method, we also directly get the results from the previous literature. Thus, we only conduct the multiple source/target domains cross-domain recognition on six possible different source- and target-domain combinations, followed \cite{Gopalan14}, among which three combinations include two source domains and one target domain, and the other three combinations include one source domain and two target domains. When there are multiple source/target domains, we simply merge the data samples in all the source/target domains as a single domain.

When there are multiple source domains, we report the results of four comparison methods, including SGF \cite{Gopalan11}, RDALR \cite{Jhuo12}, FDDL \cite{Yang11}, SDDL \cite{Shekhar13}, HMP~\cite{Bo11} and the model in \cite{Gopalan14}, as shown in Table.~\ref{tab:multiSource}. For SGF \cite{Gopalan11}, we report its performance under both unsupervised and semi-supervised settings. Note that RDALR, FDDL and SDDL are all semi-supervised methods, while our proposed method is unsupervised. It is clearly to see that the proposed method outperforms all the comparison methods significantly.

In principle, using multiple source domains should provide more information for each class, which should result in higher performance than using a single source domain. For example, for the domain combination of ``W, D$\rightarrow$A" ($41.3\%$), it shows a marginal performance improvement over the single-source domain cases, ``D$\rightarrow$A" ($38.5\%$) and ``W$\rightarrow$A"($39.1\%$). In practice, however, we do not always achieve higher performance when using multiple source domains. For example, comparing the results from Tables~\ref{tab:single} and ~\ref{tab:multiSource}, the performance of ``D, A$\rightarrow$W" ($73.2\%$) lies in between the performances of two single-source domain cases ``D$\rightarrow$W" ($89.5\%$) and ``A$\rightarrow$W"($42.4\%$). This is an interesting problem to be studied in our future work.


When there are multiple target domains, we only find two comparison methods, SGF \cite{Gopalan11} and the method from \cite{Gopalan14}. Both unsupervised and semi-supervised settings were developed for SGF, and the model from \cite{Gopalan14} is the semi-supervised based method. We take the performance from \cite{Gopalan11} and \cite{Gopalan14}, and include that in Table.~\ref{tab:multiTarget}. It can be seen that the proposed algorithm performs better than both settings of SGF and two out of three cases of model from \cite{Gopalan14}.

In principle, using multiple target domains does provide more information, since the labels are only available in the source domain. Accordingly, the performance of using multiple target domains should be the weighted (based on numbers of data samples in each involved target domain) average of the performance of using each target domain separately. The results in Tables~\ref{tab:single} and ~\ref{tab:multiTarget} are largely aligned with this expectation.

\begin{table}[htbp]
\caption{The results of multi-source domain adaptation on the object recognition dataset. Note that all the comparison methods are semi-supervised domain adaptation method except the ``US" mark one.}
\begin{center}
\renewcommand{\arraystretch}{1.3}
\def\temptablewidth{.8\columnwidth} {\rule{\temptablewidth}{1.5pt}}
\begin{tabular*}{\temptablewidth}{@{\extracolsep{\fill}}c||c|c|c}
Model                           &D, A$\rightarrow$W &A, W$\rightarrow$D &W, D$\rightarrow$A\\
\hline
SGF \cite{Gopalan11} (US)       &31.0$\pm$1.6       &25.0$\pm$0.4      &15.0$\pm$0.4 	 \\
SGF \cite{Gopalan11}            &52.0$\pm$2.5       &39.0$\pm$1.1      &28.0$\pm$0.8 	 \\
RDALR \cite{Jhuo12}             &36.9$\pm$1.1       &31.2$\pm$1.3      &20.9$\pm$0.9 	 \\
FDDL \cite{Yang11}              &41.0$\pm$2.4       &38.4$\pm$3.4      &19.0$\pm$1.2 	 \\
SDDL \cite{Shekhar13}           &57.8$\pm$2.4       &56.7$\pm$2.3      &24.1$\pm$1.6 	 \\
HMP \cite{Bo11}                 &47.2$\pm$1.9       &51.3$\pm$1.4      &37.3$\pm$1.4 	 \\
Gopalan et al.~\cite{Gopalan14} &51.3               &36.1              &35.8             \\
CJS (ours)                      &\bf{73.2$\pm$2.5} &\bf{81.3$\pm$1.3} &\bf{41.1$\pm$1.1} \\
\end{tabular*}
{\rule{\temptablewidth}{1.5pt}}
\end{center}
\label{tab:multiSource}
\end{table}

\begin{table}[htbp]
\caption{The results of multi-target domain adaptation on the 2D object recognition dataset. ``SS" and ``US" denote the semi-supervised and unsupervised setting, respectively.}
\begin{center}
\renewcommand{\arraystretch}{1.3}
\def\temptablewidth{.9\columnwidth} {\rule{\temptablewidth}{1.5pt}}
\begin{tabular*}{\temptablewidth}{@{\extracolsep{\fill}}c||c|c|c}
Model                           &W$\rightarrow$A,D &D$\rightarrow$A,W &A$\rightarrow$D,W \\
\hline
SGF \cite{Gopalan11} (US)       &28.0$\pm$1.9      &35.0$\pm$1.7      &22.0$\pm$0.2 	 \\
SGF \cite{Gopalan11} (SS)       &42.0$\pm$2.8      &46.0$\pm$2.3      &32.0$\pm$0.9 	 \\
Gopalan et al.~\cite{Gopalan14} (SS) &44.0              &\bf{49.5}         &30.0         	 \\
CJS (ours)                      &\bf{45.1$\pm$1.2} &48.4$\pm$2.2      &\bf{44.2$\pm$2.0} \\
\end{tabular*}
{\rule{\temptablewidth}{1.5pt}}
\end{center}
\label{tab:multiTarget}
\end{table}

\subsubsection{Performance under different parameter settings}\label{sec:Parameter}
There are two main parameters in the proposed algorithm: 1) the desired average size of each group constructed in the target domain using the K-means algorithm, i.e., $\gamma$; 2) the number of data samples in each anchor subspace, i.e., $N$. In order to investigate the sensitivity to different parameter settings, we tune each of the two parameters respectively, and report the performance of each parameter setting. For each parameter setting, we report the accuracy rate by percentage, for eight combinations of single source- and target- domain, and six combinations of multiple source/target domains.

\begin{figure}[htbp]
\centering
\includegraphics[width=\columnwidth]{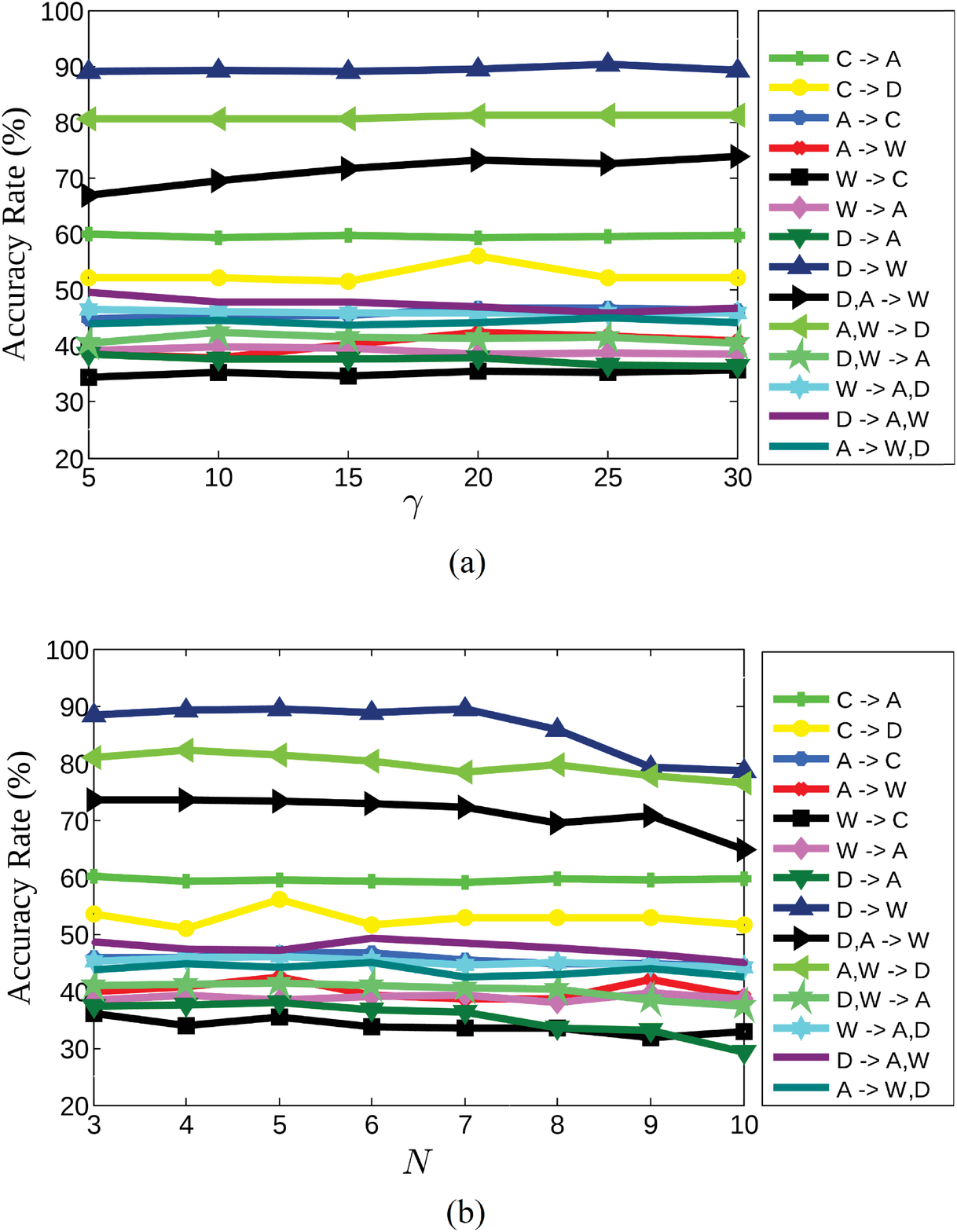}
\caption{The results when setting varies values for $\gamma$, which determines the number of groups obtained by the $K$-means algorithm. The results when setting varied values for $N$.} \label{fig:Settings}
\end{figure}

We take the value of $\gamma$ in the range of $5$ to $30$ with the step length of $5$, and the results are shown in left sub-figure in figure.~\ref{fig:Settings}. We can see that, the performance only varies in a small range for almost all the domain combinations, except for "D,A$\rightarrow$W".

For $N$, we take its value in the range of 3 to 10 with the step length of 1 and the results are reported in right sub-figure in Figure.~\ref{fig:Settings}. It is clear to see that the performance only varies in a small range for almost all the domain combinations, except for "D$\rightarrow$W" and "D,A$\rightarrow$W" when $N>8$.

Therefore, we can conclude that the performance of the proposed algorithm is not very sensitive to the two parameters $\gamma$ and $N$. In our experiments, we consistently set $\gamma$ to be 20 and $N$ to be 5.

%

\subsection{Cross-domain recognition on sentiment classification dataset}\label{sec:NLP}
Although the proposed algorithm is originally designed for vision tasks, it can be easily utilized for cross domain tasks in other areas. In this section, as an example, we compare the proposed algorithm with other 7 methods in a domain adaptation task from the natural language processing area.

In this task, customers' reviews on four different products (kitchen applications, DVDs, books and electronics) are collected as four domains \cite{Blitzer07}. Each review consists of comment texts and a rating from 0 to 5. Reviews with rating higher than 3 are classified as positive samples, and the remaining reviews are classified as negative samples. In total, there are 1000 positive reviews and 1000 negative reviews in each domain. The goal of this task is to adapt the classifier training on one domain and use it for classifying data samples in another domain.

We follow the same experiment setup described in \cite{Gong13}. In each domain, 1600 reviews including 800 positive reviews and 800 negative reviews, are used as the training set, and the rest 400 reviews are used as the testing set. We extract unigram and bigram features on the comment texts, and the feature dimension is reduced to 400. Finally, each comment text is represented by a 400-dimensional feature using the bag-of-words technique.

We conduct experiments on four pairs of source- and target- domain combinations. The same experiment has been also conducted in \cite{Gong13}. The performance is reported in Table \ref{tab:NLP_result}. It is clear to see that overall the proposed algorithm outperforms other 7 methods. From this, we can see that the proposed algorithm can also be used for domain adaption problems in non-vision areas.

\begin{table}[htbp]
\caption{Domain adaptation results on the sentiment classification. K: kitchen, D: dvd, B: books, E: electronics}
\begin{center}
\renewcommand{\arraystretch}{1.3}
\def\temptablewidth{.8\columnwidth} {\rule{\temptablewidth}{1.5pt}}
\begin{tabular*}{\temptablewidth}{@{\extracolsep{\fill}}c||c|c|c|c}
Model            &K$\rightarrow$D &D$\rightarrow$B &B$\rightarrow$E  &E$\rightarrow$K\\
\hline
TCA \cite{Pan11}        &60.4      &61.4      &61.3 	 &68.7 \\
SGF \cite{Gopalan11}    &67.9      &68.6      &66.9 	 &75.1 \\
FGK \cite{Gong12}       &69.0      &71.3      &68.4 	 &78.2 \\
SCL \cite{Blitzer06}    &72.8      &76.2      &75.0 	 &82.9 \\
KMM \cite{Huang07}      &72.2      &78.6      &76.9 	 &83.5 \\
Metric \cite{Saenko10}  &70.6      &72.0      &72.2 	 &77.1 \\
Landmark \cite{Gong13}  &75.1      &\bf{79.0} &78.5 	 &83.4 \\
CJS (ours)            &\bf{77.8} &77.0      &\bf{83.2} &\bf{84.1} \\
\end{tabular*}
{\rule{\temptablewidth}{1.5pt}}
\end{center}
\label{tab:NLP_result}
\end{table}

\section{Conclusion}\label{sec:conclusion}
This paper introduces a new subspace based domain adaptation algorithm. The compact joint subspace is independently constructed for each class, which covers both source and target domains. The compact joint subspace carries the information not only about the intrinsic characteristics of the considered class, but also about the specificity for each domain. Classifiers are trained on these compact joint subspaces. The proposed algorithm has been evaluated on two widely used datasets. Comparison results show that the proposed algorithm outperforms several existing methods on both datasets.

\ifCLASSOPTIONcaptionsoff
  \newpage
\fi


\bibliographystyle{IEEEtran}
\bibliography{activity-bib}

\begin{IEEEbiography}{Yuewei Lin} received the BS degree in optical information science and technology from Sichuan University, Chengdu, China, and the ME degree in optical engineering from Chongqing University, Chongqing, China. He is currently working toward the PhD degree in the Department of Computer Science and Engineering at the University of South Carolina. His current research interests include computer vision, machine learning and image/video processing. He is a student member of the IEEE.
\end{IEEEbiography}
\vspace{-0.5in}
\begin{IEEEbiography}{Jing Chen} received the B.S. degree in computational mathematics from Shandong University, Ji’nan, China, the M.S. and Ph.D. degrees in computer science and technology from Chongqing University, Chongqing, China. He is currently a postdoctoral fellow with the Faculty of Science and Technology, University of Macau, Macau. His research interests include machine learning, pattern recognition, and biomedical information processing and analysis.
\end{IEEEbiography}
\vspace{-0.5in}
\begin{IEEEbiography}{Yu Cao} received the B.S. degree in Information and Computation Science from Northeastern University, Shenyang, China, 2003; and received the M.S. degree in Applied Mathematics from Northeastern University, Shenyang, China, 2007. He received Ph.D. degree in Computer Science and Engineering from University of south Carolina, USA, 2013. He is currently a Post-doctoral Researcher in IBM Almaden Research Center, San Jose, CA. His research interests include computer vision, machine learning, pattern recognition, medical image processing. He is a member of IEEE.
\end{IEEEbiography}
\vspace{-0.5in}
\begin{IEEEbiography}{Youjie Zhou} is currently pursuing the Ph.D. degree in computer science and engineering with the University of South Carolina, Columbia, SC, USA, as a Research Assistant with the Computer Vision Laboratory. His current research interests include computer vision, machine learning, and large-scale multimedia analysis. He received the B.S. degree in software engineering from East China Normal University (ECNU), Shanghai, China, in 2010. From 2007 to 2010, he was a Research Assistant with the Institute of Massive Computing, ECNU, where he worked on multimedia news exploration and retrieval. He is a student member of the IEEE.
\end{IEEEbiography}
\vspace{-0.5in}
\begin{IEEEbiography}{Lingfeng Zhang} received the mathematics BS degree and the computer science MS degree in Chongqing University, P. R. China in 2009 and 2012 respectively. He is currently working toward the PhD degree in University of Houston. He is interested in machine learning and big data analysis. He is a student member of the IEEE. 
\end{IEEEbiography}
\vspace{-0.5in}
\begin{IEEEbiography}{Yuan Yan Tang (F'04)} is a Chair Professor in Faculty of Science and Technology at University of Macau and Professor/Adjunct Professor/Honorary Professor at several institutes in China, USA, Canada, France, and Hong Kong. His current interests include wavelets, pattern recognition, image processing, and artificial intelligence. He has published more than 400 academic papers and is the author/coauthor of over 25 monographs/books/bookchapters. He is the Founder and Editor-in-Chief of International Journal on Wavelets, Multiresolution, and Information Processing (IJWMIP), and Associate Editors of several international journals. He is the Founder and Chair of pattern recognition committee in IEEE SMC. He has serviced as general chair, program chair, and committee member for many international conferences. Dr. Tang is the Founder and General Chair of the series International Conferences on Wavelets Analysis and Pattern Recognition (ICWAPRs). He is the Founder and Chair of the Macau Branch of International Associate of Pattern Recognition (IAPR). Dr. Y. Y. Tang is a Fellow of IEEE, and Fellow of IAPR.
\end{IEEEbiography}
\vspace{-0.5in}
\begin{IEEEbiography}{Song Wang} received the Ph.D. degree in electrical and computer engineering from the University of Illinois at Urbana-Champaign (UIUC), Urbana, IL, USA, in 2002. From 1998 to 2002, he was a Research Assistant with the Image Formation and Processing Group, Beckman Institute, UIUC. In 2002, he joined the Department of Computer Science and Engineering, University of South Carolina, Columbia, SC, USA, where he is currently a Professor. His current research interests include computer vision, medical image processing, and machine learning. He is currently serving as the Publicity/Web Portal Chair of the Technical Committee of Pattern Analysis and Machine Intelligence, the IEEE Computer Society, and an Associate Editor of Pattern Recognition Letters. He is a member of the IEEE Computer Society.
\end{IEEEbiography}






\end{document}